\documentclass[10pt,twocolumn,letterpaper]{article}

\usepackage[pagenumbers]{cvpr} % To force page numbers, e.g. for an arXiv version

\usepackage[utf8]{inputenc} % allow utf-8 input
\usepackage[T1]{fontenc}    % use 8-bit T1 fonts
\usepackage{booktabs}       % professional-quality tables
\usepackage{amsfonts}       % blackboard math symbols
\usepackage{nicefrac}       % compact symbols for 1/2, etc.
\usepackage{microtype}      % microtypography
\usepackage{amsmath}
\usepackage{subcaption}
\usepackage{float}
\usepackage{adjustbox}
\usepackage{multirow}
\usepackage{makecell} 
\usepackage[accsupp]{axessibility}  % Improves PDF readability for those with disabilities.
\usepackage{xspace}
\usepackage{wrapfig}
\usepackage{colortbl}
\usepackage{graphicx}
\usepackage{natbib}
\definecolor{cvprblue}{rgb}{0.21,0.49,0.74}
\usepackage[pagebackref,breaklinks,colorlinks,allcolors=cvprblue]{hyperref}

\title{\method Boost Spatial-Temporal Reasoning \\in Multimodal Language Model}

\author{\textbf{Benlin Liu\textsuperscript{1}\thanks{Equal contribution.}, Yuhao Dong\textsuperscript{2,3}$^{\star}$, Yiqin Wang\textsuperscript{2}$^{\star}$, }\\
\textbf{Zixian Ma\textsuperscript{1}, Yansong Tang\textsuperscript{2}, Luming Tang\textsuperscript{4}, Yongming Rao\textsuperscript{3},  Wei-Chiu Ma\textsuperscript{5,6}, Ranjay Krishna\textsuperscript{1,5}}\\ 
\textsuperscript{1}University of Washington,  \textsuperscript{2}Tsinghua University, 
\textsuperscript{3}Tencent, \\
\textsuperscript{4}Google Deepmind,
\textsuperscript{5}Allen Institute for AI,
\textsuperscript{6}Cornell University \\
\href{https://coarse-correspondence.github.io}{coarse-correspondence.github.io}
}

\newcommand{\method}{\textsc{Coarse Correspondences}\xspace}
\newcommand{\methodtable}{\textbf{\textit{Coarse Correspondences}}\xspace}

\newcommand{\PreserveBackslash}[1]{\let\temp=\\#1\let\\=\temp}
\newcolumntype{C}[1]{>{\PreserveBackslash\centering}p{#1}}
\newcolumntype{L}[1]{>{\PreserveBackslash\raggedright}p{#1}}

\begin{document}
\maketitle
\begin{abstract}
Multimodal language models (MLLMs) are increasingly being applied in real-world environments, necessitating their ability to interpret 3D spaces and comprehend temporal dynamics. 
Current methods often rely on specialized architectural designs or task-specific fine-tuning to achieve this. We introduce \method, a simple lightweight method that enhances MLLMs' spatial-temporal reasoning with 2D images as input, without modifying the architecture or requiring task-specific fine-tuning.
Our method uses a lightweight tracking model to identify primary object correspondences between frames in a video or across different image viewpoints, and then conveys this information to MLLMs through visual prompting. 
We demonstrate that this simple training-free approach brings substantial gains to GPT4-V/O consistently on four  benchmarks that require spatial-temporal reasoning, including \textbf{+20.5\%} improvement on ScanQA, \textbf{+9.7\%} on OpenEQA's episodic memory subset, \textbf{+6.0\%} on the long-form video benchmark EgoSchema, and \textbf{ +11\%} on the R2R navigation benchmark. 
Additionally, we show that \method can also enhance open-source MLLMs' spatial reasoning (by \textbf{+6.9\%} on ScanQA) when applied in both training and inference and that the improvement can generalize to unseen datasets such as SQA3D (\textbf{+3.1\%}).
Taken together, we show that \method  effectively and efficiently boosts models' performance on downstream tasks requiring spatial-temporal reasoning.
\vspace{-7pt}
\end{abstract}    
\section{Introduction}
\label{sec:intro}

Intelligence is multi-faceted. While multi-modal language models~\cite{openai2024gpt-4o} have shown remarkable linguistic, logical and even mathematical intelligence, many remain doubtful about their visual and spatial intelligence. Despite their excellent performance on visual-lingusitic tasks, many recent works~\cite{OpenEQA2023,singh2024evaluating} demonstrate that state-of-the-art MLLMs still struggle at 3D and long video understanding benchmarks, performing only marginally better than blind text-only baselines. These results suggest that spatial-temporal reasoning is a major bottleneck on MLLMs' path to general visual intelligence.

To enhance MLLMs' 3D understanding, researchers have mainly explored three approaches: providing MLLMs with 3D data as input~\cite{xu2025pointllm}, designing specialized architectures for 3D tasks~\cite{3d-llm}, or employing supervised fine-tuning with 3D data~\cite{chen2024spatialvlm}.
Similarly, to boost MLLMs' temporal understanding, prior works have proposed new model architectures designed for long video understanding~\cite{papalampidi2023simple,balažević2024memory}, or adopted Socratic-based methods~\cite{zhang2024simple,kahatapitiya2024language} (i.e., converting each frame of a video into text using a caption model, and then using text-only LLMs  to summarize). 

\begin{figure*}[t]
    \centering
    \includegraphics[width=\linewidth]{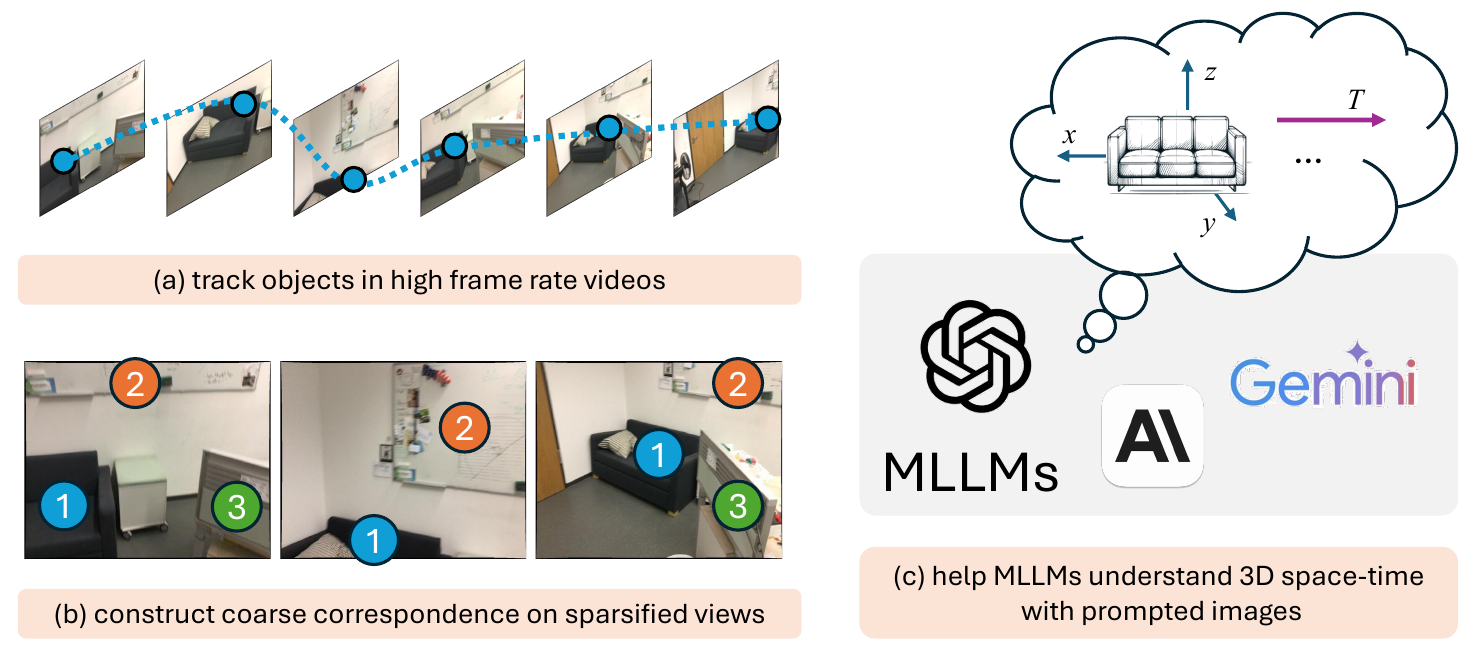} 
    \vspace{-12pt}
    \caption{We combined light-weight video tracking models and multimodal LLMs to achieve a better understanding of 3D spacetime. (a) We use a tracking model at a high frame rate to obtain instance segmentation masks for each frame. (b) Then, we sequentially sparsify input frames, select prominent coarse correspondences, and visualize the constructed coarse correspondences on the images. (c) Finally, we enable MLLMs to better understand 3D spacetime from the prompted images.}
    \vspace{-10pt}
    \label{fig:pipeline}
\end{figure*}

In contrast to prior works that separately enhance spatial and temporal reasoning, we propose \method, a simple yet effective training-free visual prompting method to jointly boost spatial-temporal reasoning in MLLMs.
\method uses a tracking model to extract object-level correspondences across multiple images, and then represent the most salient correspondence relationships on the images through visual prompting. 
We refer to our method as \textit{Coarse} because of the following: 
\begin{enumerate}
    \item We only visually prompt for instance-level correspondences and not point-level correspondences.
    \item The instance-level correspondences are extracted using off-the-shelf tracking models. 
    \item We only visualize a handful of prominent corresponding instances.
\end{enumerate}
Despite the obtained correspondence information not being perfectly precise, our method can still significantly boost MLLMs' spatial-temporal reasoning with only 2D image inputs and without any specialized architectural design or task-specific fine-tuning.

We have demonstrated substantial performance gains of \method through extensive experiments with both open-source and closed-source models across 6 benchmarks on spatial-temporal reasoning.
For closed-source models, we apply \method on GPT4-V/O during inference and achieve compelling gains.  
First, on spatial reasoning, we show that our method significantly surpasses state-of-the-art models by 20.5\% and 9.7\% on ScanQA~\cite{azuma2022scanqa} and OpenEQA~\cite{OpenEQA2023} respectively. Second, for long video understanding, our method leads to a 6\% gain in performance on the EgoSchema benchmark~\cite{mangalam2023egoschema}. 
Notably, our method uses much fewer input images and, in a zero-shot manner, outperforms many fine-tuned models that use far more images. For example, on EgoSchema, \method surpasses state-of-the-art results with just 8 uniformly sampled frames from a 3-minute video, greatly reducing the computational costs of MLLMs compared to existing methods. 
In addition to 3D and video QA tasks, we further demonstrate that our method enhances models' performance on embodied tasks such as navigation~\cite{krantz2020beyond}, which require strong spatial-temporal reasoning, by 11\% in success rate on R2R. 
These results suggest that \method boosts MLLMs' spatial-temporal reasoning both effectively and efficiently. 
Last but not least, we experiment with open-source MLLMs~\cite{liu2024llava} by applying \method in both instruction tuning and inference; again, our method shows performance gains against the baseline (by 6.9\% on ScanQA), and the improvement even generalizes to unseen datasets such as SQA3D (+3.1\%). These results suggest that \method works well universally with any model -- both closed-source and open-source -- that can take in multiple images and understand visual markers. 

Overall, we want to highlight with this work that, despite its simplicity and being underestimated for semantic tasks in deep learning, visual correspondence can still bring significant utility to spatial-temporal reasoning in MLLMs, just as it has long contributed to 3D reconstruction~\cite{schonberger2016structure}.
We hope our work demonstrates the potential of leveraging general-purpose MLLMs to better understand our physical world.
\vspace{-10pt}
\section{Method}
We introduce \method, a visual prompting method that allows MLLMs to reason about 3D space and time. 

\noindent \textbf{Problem formulation.} Given a question $\mathcal{Q}$ and a sequence or set of observations in an environment $[I_1, \ldots, I_n]$, 
our aim is to design a visual prompt $\mathcal{P}(\ldots)$ that modifies the input image set. These image inputs don't have to be a video. They can also represent a set of images of a specific scene from multiple viewpoints.
We evaluate the prompt by measuring its utility in prompting an MLLM $\mathcal{M}$:
\begin{align*}
    & [I_1^\prime, \dots, I_n^\prime] = P([I_1, \dots, I_n])\\
    & \hat{\mathcal{A}} = \mathcal{M}(([I_1^\prime, \dots, I_n^\prime]), \mathcal{Q})
\end{align*}
We compare generated answer $\hat{\mathcal{A}}$ with the ground truth $\mathcal{A}$.

In our framework, the MLLMs can be any general-purpose model without requiring any special architecture or training for spatial-temporal reasoning. Our aim is to develop a prompting strategy that allows models to improve such capabilities without any training (Figure~\ref{fig:pipeline}).

\subsection*{Coarse Correspondence}
Our prompting method, \method, contains four steps: (1) tracking correspondences, (2) sparsify frames, (3) selecting, and (4) visualizing coarse correspondences. 

\noindent \textbf{(1) Tracking correspondences.} Given $n$ input images, $[I_1, \ldots, I_n]$, we first use an off-the-shelf video object tracking model, such as Tracking Anything~\citep{yang2023track}. This model extracts class-agnostic instance segmentation masks $(M_1, \ldots, M_n )$ for each image. Each $M_i$ is a $H \times W$ dimensional matrix where $H$ and $W$ are the height and width of the input image $I_i$. Each pixel location in $M_i$ contains an instance ID, indicating which instance the pixel at that position belongs to within the image sequence.

\noindent \textbf{(2) Sparsify frames.}
Since most MLLMs contain a large number of parameters, directly using them to process long image sequences is very computationally intensive. Additionally, proprietary MLLMs like GPT-4O can also incur significant costs if the number of image tokens that need to be processed increases. Reducing the number of input images might lose vital information necessary for MLLMs. 

\method strikes a balance in this tradeoff by extracting meaningful video object tracks (a relatively cheaper operation) from high-frame-rate image sequences, and then samples a few image inputs along with the tracks, to retain—and even improve—performance while reducing the MLLM's computation cost.
From this extracted video object tracks, we perform temporal downsampling, retaining only $m << n$ uniformly sampled images and their corresponding masks, denoted as $[I_{s_1}, \ldots, I_{s_m}]$ and $[M_{s_1}, \ldots, M_{s_m}]$, where $s_i \in \{1, \dots, n\}$.
This downsampling reduces the number of images we feed into $\mathcal{M}$. 
%More complex sampling can be considered as one future directions.

\noindent \textbf{(3) Selecting coarse correspondences.}
Prompting an MLLM with all the detected correspondences results in information overload. In fact, our ablations (discussed in Sec~\ref{sec:analysis}) find that adding all the correspondences reduces the MLLM's performance.
Therefore, we select a subset of prominent instances to retain. We select the prominent instances of the top-K objects that co-occur in the most number of frames.
We first calculate the occurrence frequency and area sum of each unique instance ID in the retained $m$ masks using the following equation:
\begin{align*}
\mathcal{F}req(\text{ID}) &= \sum_{i=s_1}^{s_m} \mathbf{1}_{\{\text{ID} \in M_i\}},
\mathcal{A}rea(\text{ID}) &= \sum_{i=s_1}^{s_m} \sum_{p \in M_i} \mathbf{1}_{\{\text{ID} = p\}}.
\end{align*}
Then, we first sort all instance IDs in descending order based on $\mathcal{F}req(\text{ID})$. If there are ties, we further sort based on $\mathcal{A}rea(\text{ID})$. Finally, we retain the top $k$ instance IDs as tracklets, denoted as $[T_1, \dots, T_k]$, to visualize for MLLMs.

\noindent \textbf{(4) Visualizing coarse correspondences.}
For each set of obtained correspondence relationships, we visualize the correspondences directly in the image as a marker. Specifically, for each identified primary instance ID $T_i$, if it exists in the mask $M_{s_j}$ of a retained image $I_{s_j}$, we overlay a mark with a fixed size and shape labeled with $T_i$ at the position $(\bar{x}_{ij}, \bar{y}_{ij})$ on $I_{s_j}$ to produce $I_{s_j}^\prime$. The specific placement position can be easily obtained by:
\begin{align*}
(\bar{x}_{ij}, \bar{y}_{ij}) = \frac{\sum_{(x,y)} (x, y) \cdot \mathbf{1}_{\{M_{s_j}(x,y) = T_i\}}}{\sum_{(x,y)} \mathbf{1}_{\{M_{s_j}(x,y) = T_i\}}}
\end{align*}

Naturally, we can overlay not just the markers but also the segmentation outlines or even the segmentation masks associated with each retained prominent instance. We explore these ablations later. In the end, we obtain the prompted image sequence $[I_1^\prime, \ldots, I_m^\prime]$, which is then used as the input to MLLMs.

\section{Prompting Proprietary Models}

We first evaluated the utility of our \method on multiple tasks using proprietary models, including understanding 3D space (ScanQA~\citep{azuma2022scanqa} and OpenEQA~\citep{OpenEQA2023} in \S\ref{sec:spatial_experiments}) as well as temporal events (EgoSchema~\citep{mangalam2023egoschema} in \S\ref{sec:temporal_experiments}). Building on the improvements our method brings to spatial-temporal reasoning, we further demonstrate that our method also delivers significant gains in navigation tasks (VLN-CE~\citep{krantz2020beyond}). 
Across all these benchmarks, we augment proprietary MLLMs (e.g., GPT-4V and GPT-4O) with \method and evaluate its zero-shot performance. We show that \method significantly improves the base GPT models and can substantially surpass many current state-of-the-art methods that have undergone specialized fine-tuning, even while using much fewer images as input. All experiments were conducted using A100 80G GPUs.

\begin{table*}[t]
  \centering
    \adjustbox{width=\linewidth}{
    \begin{tabular}{L{200pt}C{20pt}C{35pt}C{35pt}C{35pt}C{50pt}C{35pt}}
    \toprule
    Model & Frame & BLEU-1 & BLEU-2 & METEOR & ROUGE-L & CIDEr\\
    \midrule
    \textit{3D-Specific Models} \\
    \midrule
    ScanQA~\citep{azuma2022scanqa} & - &  26.9 & 16.6 & 11.5 & 30 & 55.4\\
    ScanRefer+MCAN~\citep{yu2019deep} & - & 30.2 & 20.4 
 & 13.1 & 33.3 & 64.9\\
    3D-LLM~\citep{3d-llm} & - & 39.3 & 25.2 & 14.5 & 35.7 & 69.4\\
    \midrule
   %  Gemini & 8 & 24.1 & 13.5& 11.3& 35.4 & 68.3 \\
   % \rowcolor{gray!20}Gemini+\methodtable & 8 & 25.4 & 15.7 & 12.0 & 37.1 & 75.5 \\
   %  Claude & 8 & 19.8 & 11.1& 10.0& 29.3 & 57.7 \\
   % \rowcolor{gray!20}Claude+\methodtable & 8 & 27.1 & 23.9 & 11.7& 33.1& 65.7 \\
   \textit{Open-source Multi-modal Models} \\
   \midrule
   LLaVA(Fine-tuned) & 64 & 34.7 & 22.0 & 13.8 & 31.1 & 67.3 \\
   \rowcolor{gray!20}LLaVA+\methodtable & 64 & \textbf{38.6} & \textbf{24.7} & \textbf{15.4} & \textbf{38.3} & \textbf{74.2} \\
   \midrule
   \textit{Proprietary Multi-modal Models} \\
   \midrule
    GPT-4V & 8 & 28.6 & 13.4& 13.5& 33.4 & 59.6 \\
    \rowcolor{gray!20} GPT-4V+\methodtable & 8 & \textbf{39.7} & \textbf{25.5} & \textbf{17.4} & \textbf{40.8} & \textbf{79.2} \\
    \midrule
    GPT-4O & 4 & 30.5 & 19.8 & 14.8 & 36.1 & 72.2 \\
    \rowcolor{gray!20} GPT-4O+\methodtable & 4 & \textbf{35.4} & \textbf{25.5} & \textbf{18.0}& \textbf{42.6} & \textbf{87.0} \\
    \bottomrule
    \end{tabular}%
    }
    \vspace{-8pt}
    \caption{\textbf{Comparison on ScanQA validation set.} We conduct experiments on the ScanQA validation set to demonstrate the effectiveness of \method with different MLLMs. Our method enables both proprietary models and open-source models to surpass all 3D-specific models.}
  \label{tab:sota}
  \vspace{-15pt}
\end{table*}%

\subsection{Spatial understanding}
\label{sec:spatial_experiments}

\noindent \textbf{Benchmarks}.
The validation set of ScanQA dataset contains 4675 questions about 71 scenes. 
The questions % in ScanQA 
require basic recognition, 3D localization, and 3D embodied capabilities~\citep{duan2022survey}. 
%The validation set 
It contains two ground-truth answers per question for evaluation with models that produce free-form answers.
OpenEQA Dataset is an open-vocabulary dataset benchmarking spatial environment understanding and embodied reasoning. 
We evaluate on OpenEQA's EM-EQA data split, which contains over 1600 high-quality human-generated questions.
The subset tests the episodic memory of an agent moving through a 3D environment over time.

\noindent \textbf{Baselines.} For ScanQA, we evaluate \method by augmenting both GPT-4\{V,O\}, Gemini and Claude models. 
Besides, we also consider 3D specialized models~\citep{yu2019deep, azuma2022scanqa, 3d-llm} fine-tuned on ScanQA. 
For OpenEQA, we compare against language-only models to account for language bias (LLaMA2~\citep{touvron2023llama}), commonly used general-purpose multimodal LLMs (GPT-4~\citep{openai2024gpt4}, Claude3~\citep{anthropic2024claude}, Gemini-Pro~\citep{geminiteam2024gemini}, GPT-4V with 15 and 50 frames.

\noindent \textbf{Metrics.} For ScanQA, following prior works, we adopt BLEU~\citep{papineni2002bleu} %scores
, METEOR~\citep{banerjee2005meteor}, ROUGE-L~\citep{lin2004rouge}, and CIDEr~\citep{vedantam2015cider} as our evaluation metrics. For OpenEQA, we follow their evaluation approach by using GPT-4 to compare the generated answers with the ground-truth answers and assign a score. We report the average score across all questions.

\noindent \textbf{Results.}
For ScanQA, as shown in Table~\ref{tab:sota}, compared to raw input, \method consistently improves the overall performance of different proprietary models. For instance, on the strongest model, GPT-4o, \method brings improvements of $5.7$ BLEU-2, $3.2$ METEOR, $6.5$ ROUGE-L, and $15$ CIDEr points. Compared to methods that are specifically designed for 3D understanding tasks, fine-tuned with specialized 3D SFT data, or even those that use 3D point clouds instead of 2D images as input, we observe that a general-purpose MLLM can still outperform them, especially when enhanced with \method. Moreover, we found that this can be achieved using far fewer images as input. 

We also demonstrated the same conclusion on OpenEQA, as indicated in Table~\ref{tab:openeqa}. By applying \method, we significantly improved the performance of both GPT-4v and GPT-4o, achieving better results with fewer input images. These findings suggest that general-purpose MLLMs are indeed capable of understanding 3D space, and \method can significantly enhance their spatial understanding while simultaneously reducing the number of views needed, which could lower the inference cost and make MLLMs more useful for embodied tasks.

\begin{table}[ht]
  \centering
  \small
  \adjustbox{width=.95\linewidth}{
    \begin{tabular}{C{120pt}|C{20pt}|C{30pt}}
      \toprule
      Models & Frame & Accuracy \\
      \midrule
      LLaMA2~\citep{touvron2023llama} & 0 & 28.3 \\
      GPT-4~\citep{openai2024gpt4} & 0 & 33.5 \\
      Claude3~\citep{anthropic2024claude} & 20 & 36.3 \\
      Gemini-Pro~\citep{geminiteam2024gemini} & 15 & 44.9 \\
      GPT-4V~\citep{OpenAI_GPT4V} & 15 & 54.6 \\
      GPT-4V~\citep{OpenAI_GPT4V} & 50 & 55.3 \\
      Human & Full & 86.8 \\
      \midrule
      GPT-4V & 8 & 44.8 \\
      \rowcolor{gray!20} GPT-4V+CC & 8 & \textbf{58.5} \\
      \midrule
      GPT-4O & 4 & 49.4 \\
      \rowcolor{gray!20} GPT-4O+CC & 4 & \textbf{59.1} \\
      \bottomrule
    \end{tabular}
  }
  \vspace{-8pt}
  \caption{\textbf{Comparisons on EM-EQA setting of OpenEQA.} Our method further enhances the embodied ability of MLLMs and exceeds previous methods by a large margin.}
  \label{tab:openeqa}
  \vspace{-10pt}
\end{table}

\begin{table}[ht]
  \centering
  \small
  \adjustbox{width=.95\linewidth}{
    \begin{tabular}{C{120pt}|C{20pt}|C{30pt}}
      \toprule
      Models & Frame & Accuracy \\
      \midrule
      LongViviT~\citep{papalampidi2023simple} & 256 & 56.8 \\
      MC-ViT-L~\citep{balažević2024memory} & 128+ & 62.6 \\
      LLoVi~\citep{zhang2024simple} & 180 & 58.3 \\
      VideoAgent~\citep{wang2024videoagent} & 8.4 & 60.2 \\
      MVU~\citep{ranasinghe2024understanding} & 16 & 60.3 \\
      VideoAgent~\citep{fan2024videoagent} & - & 62.8 \\
      LangRepo~\citep{kahatapitiya2024language} & - & 66.2 \\
      \midrule
      GPT-4V & 8 & 64.2 \\
      \rowcolor{gray!20} GPT-4V+CC & 8 & \textbf{67.4} \\
      \midrule
      GPT-4O & 8 & 67.2 \\
      \rowcolor{gray!20} GPT-4O+CC & 8 & \textbf{73.2} \\
      \bottomrule
    \end{tabular}
  }
  \vspace{-8pt}
  \caption{\textbf{Comparisons on EgoSchema validation set.} \method improves existing MLLMs and surpasses previous fine-tuned models in a zero-shot manner.}
  \label{tab:egoschema}
  \vspace{-10pt}
\end{table}

\subsection{Temporal understanding}
\label{sec:temporal_experiments}

\noindent \textbf{Benchmarks.} We evaluated the improvements of our method for long video understanding using the EgoSchema dataset. Each video in EgoSchema is 3 minutes long, with a corresponding question that includes five multiple-choice options. These questions are designed to ensure that answering them requires viewing a sufficient number of frames from the video. Due to budget constraints, we limited our evaluation to 500 questions from the validation set. 

\noindent \textbf{Baselines.}
The baseline methods we compared against include newly designed and trained model architectures specifically for long video understanding, such as LongViviT~\citep{papalampidi2023simple} and MC-ViT-L~\citep{balažević2024memory}. On the other hand, we also compared methods that rely on text-only foundation models (e.g., GPT-4), i.e., Socratic-based approaches~\citep{zhang2024simple,kahatapitiya2024language}, which first use an off-the-shelf image captioning model~\citep{zhao2023learning} to convert video frames into captions, and then prompt GPT-4 to answer questions based on those captions. Additionally, we compared agent-based methods~\citep{wang2024videoagent,fan2024videoagent}, which involve using GPT-4 alongside an image captioning model in an agent framework to iteratively perform a series of multi-step reasoning operations to understand long videos. 
In contrast to these approaches, our method is entirely based on an end-to-end general MLLM architecture, exploring how to further enhance its ability to understand long videos.

\noindent \textbf{Results.} 
\method demonstrates state-of-the-art performance, significantly outperforming existing approaches in a zero-shot manner (Table~\ref{tab:egoschema}). Compared to the original GPT-4o model, our method improves its performance by 6$\%$. Notably, our method uses far fewer frames than other approaches, yet achieves higher results compared to methods that use many more frames. It is also worth highlighting that even the original GPT-4o, when limited to just 8 frames, already serves as a very strong baseline. This indicates the potential of a sufficiently powerful general-purpose MLLM in long video understanding. We provide a detailed breakdown of \method's performance across different question types in the supplementary material.

\begin{table*}[h]
  \small\centering
    \adjustbox{width=\linewidth}{
        \begin{tabular}
        {C{50pt}|C{60pt}C{60pt}C{70pt}C{60pt}C{60pt}}
        \toprule 
        Methods & \makecell{Success \\ Rate $\uparrow$} & \makecell{Oracle \\ Success Rate $\uparrow$} & \makecell{Success weighted \\ by Path Length $\uparrow$} & \makecell{Trajectory \\ Length $\uparrow$} & \makecell{Navigation \\ Error $\downarrow$} \\
       \midrule

        GPT-4O  & 12.00 & 18.00 & 10.37 & 7.31 & 8.49 \\
        \rowcolor{gray!20} GPT-4O+CC & \textbf{23.00} & \textbf{29.00} & \textbf{21.03} & \textbf{8.12} & \textbf{7.37} \\
        \bottomrule
        \end{tabular}
    }
    \vspace{-8pt}
    \caption{\textbf{Comparison on Navigation task.} \method improve GPT-4o's performance on R2R dataset for different evaluation metrics. 
    %Except for NE, where a lower value indicates better performance, higher values for the other metrics reflect better performance.
    }
  \label{tab:nav}%
  \vspace{-7pt}
\end{table*}%

\subsection{Navigation}
\label{sec:naviagtion}
Building on the improvements in 3DQA and VideoQA, we hope that our method can also prove effective in embodied tasks such as navigation. Navigation requires an agent to understand 3D space, such as being able to determine the spatial relationship between objects in the instruction and itself, while also performing temporal reasoning to assess the progress toward completing the instruction. We consider conducting experiments on the VLN-CE benchmark~\citep{krantz2020beyond}, which is a continuous simulation environment for low-level action execution in indoor scenes.

\noindent \textbf{Setup.} 
We adopt the val-unseen split from R2R~\citep{krantz2020beyond} for evaluation. Unlike the previous QA tasks, where all images could be processed at once, in navigation tasks, each image is processed in an online fashion. Specifically, we feed in one image at each iteration of the conversation. Given the significant variation in viewpoints during navigation, we use SAMv2~\citep{ravi2024sam}, the state-of-the-art model for long-range tracking, to label each new input image based on episodic history. Then, using the prompted images, we induce the MLLM to output one of four actions at each step: FORWARD (distance), TURN-LEFT (rotation angle), TURN-RIGHT (rotation angle), and STOP. We follow NavGPT~\citep{zhou2024navgpt} to craft input prompts. Considering the high computational cost of navigation tasks, we selected 100 samples from the val-unseen split. Our primary goal is to demonstrate that our method can enhance GPT models' capabilities in zero-shot navigation tasks, which remains a significant challenge for various types of models.

\noindent \textbf{Metrics.} 
We follow the standard VLN evaluation metrics  to evaluate the navigation performance, including success rate (SR), oracle success rate (OS), success weighted by path length (SPL), trajectory length (TL), and navigation error from goal (NE). Note that an episode is considered successful if the agent calls the STOP action within 3m of the goal in the VLN-CE. 

\noindent \textbf{Results.} 
As shown in Table~\ref{tab:nav}, our method achieved improvements across all metrics.  
While GPT-4o performs impressively on many QA tasks, its zero-shot performance on navigation tasks is relatively low. This may partly be due to the lack of specialized training on action data, making it less accurate in outputs such as determining how many meters to move forward. However, our method reveals another dimension of the problem: MLLMs’ understanding of the 3D spacetime in which they operate can be further enhanced. This is evidenced by the significant improvements in navigation when using \method. We believe that our approach holds great potential for embodied tasks, which can be explored in future research.

\section{Prompting Open Models}
We further validate the effectiveness of our \method on open-source models. Our primary goal is to demonstrate that our method not only enhances proprietary models but also yields general improvements across various MLLMs. However, directly prompting open-source models presents two challenges: first, these models lack the ability to interpret prompting marks as effectively as proprietary models; second, they currently cannot handle multiple image inputs well. Thus, we first fine-tune the open-source models to understand prompting marks and manage multiple image inputs, enabling \method to enhance these models’ environmental understanding capabilities at the inference stage. In this work, we primarily focus on proving that \method improves open-source models' understanding of 3D space.

We start with the LLaVA model~\citep{liu2024llava} and fine-tune it using a dataset comprising approximately 1.2 million samples of image and video data. Notably, the ScanQA dataset is the \textit{only} 3D-related dataset in this collection. \method is applied exclusively to the ScanQA data, while other data maintain their original format. Fine-tuning on such a dataset enables the open-source LLaVA model to better interpret overlayed marks on images and process multiple image inputs.

\textbf{In-domain Evaluation.} We first evaluate whether \method can improve open-source models on the ScanQA validation set. As shown in Table~\ref{tab:sota}, our method significantly enhances the model's 3D spatial understanding compared to the baseline fine-tuned on the same data without using \method, even outperforming previous VLMs~\cite{3d-llm} specifically designed for 3D tasks, which incorporate specialized architectural designs and are fine-tuned on substantially larger amounts of 3D-related data. This in-domain evaluation demonstrates that \method is also effective for open MLLMs.
\begin{table}[t]
  \centering
  \begin{tabular}{C{180pt}C{30pt}}
    \toprule
    Method & Acc\\
    \midrule
    LLaVA(w/o CC during finetune) & 36.0\\
    LLaVA+CC(train only) & 37.17\\
    LLaVA+CC(train \& test) & 39.13\\
    \bottomrule 
  \end{tabular}
  \vspace{-7pt}
  \caption{\textbf{Comparisons on SQA3D dataset.} We tested the effectiveness of \method on the out-of-domain SQA3D dataset. With the same training data, we found that using \method not only improves performance during inference but also enhances zero-shot evaluation when applied solely during the training phase.}
  \label{tab:sqa3d}
  \vspace{-10pt}
\end{table}

\textbf{Out-domain Evaluation.} To further demonstrate the generalizability of our method, we conduct experiments on the SQA3D dataset to evaluate whether \method can also provide improvements in a zero-shot setting. As shown in Table~\ref{tab:sqa3d}, on this previously unseen dataset, \method still outperforms the baseline simply fine-tuned on the same data, proving that our method can enhance open-source models' understanding of 3D space on out-of-domain datasets.

Notably, even without using \method during inference, simply applying it during training already yields improvements. This highlights that our method is effective not only as a prompting technique for inference but also holds potential as a data augmentation method during training, a prospect worth exploring further in future work.

\section{Analysis}
\label{sec:analysis}

Here, we explore the various design decisions in our method. Although the study is conducted only on ScanQA, applying the design choices derived from ScanQA to all other benchmarks also yields significant performance improvements. This strongly demonstrates the generalizability of these design choices. Additionally, we further analyze the impact of our method on mitigating the camera motion bias present in MLLMs.

\begin{figure*}
    \centering
    \includegraphics[width=\linewidth]{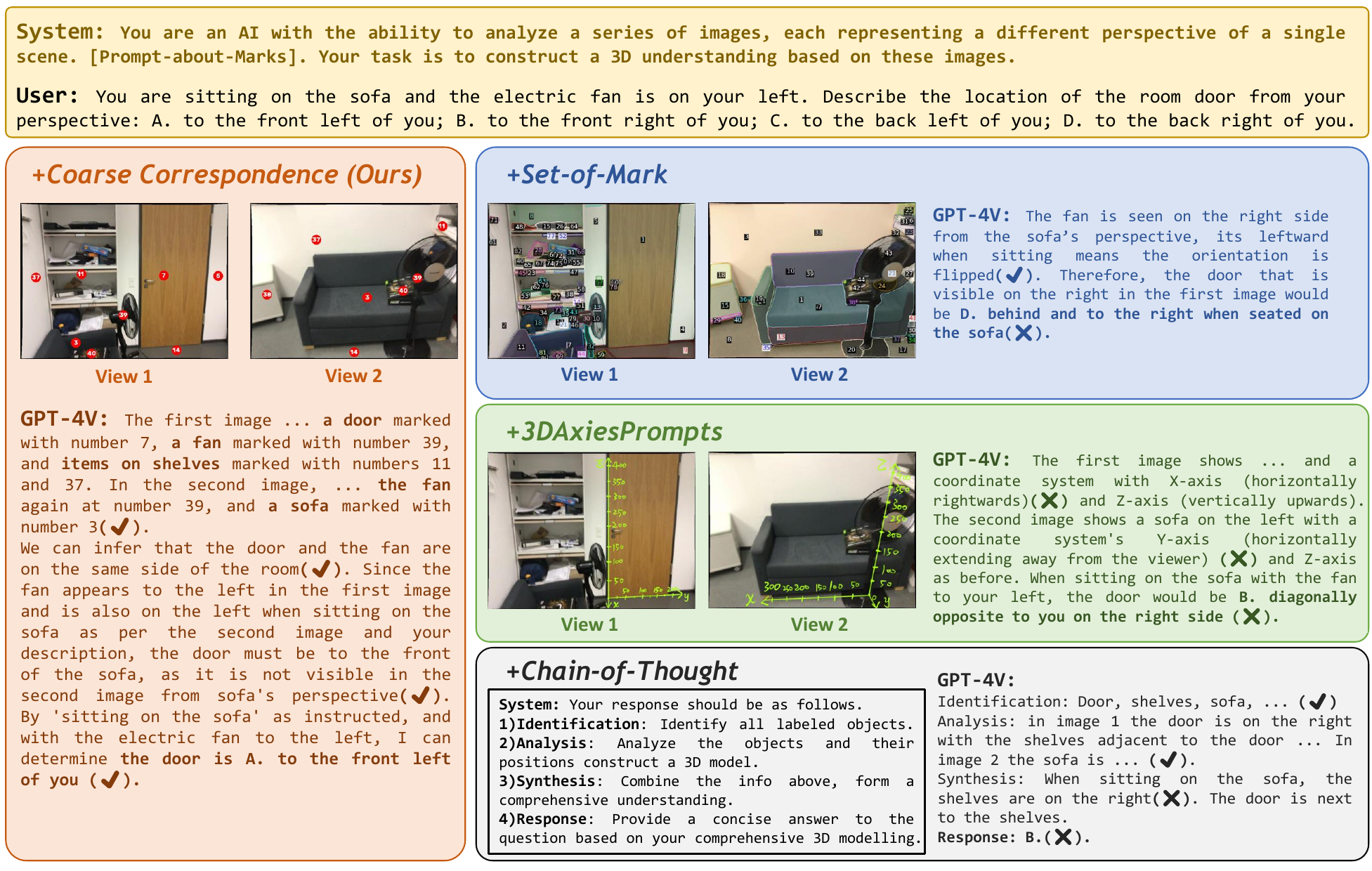}
    %\vspace{-20pt}
    \caption{\textbf{Comparison of different prompting method.} 
    Our proposed \textcolor{orange}{Coarse Correspondences} 
    successfully guided GPT-4O to understand 3D spatial relationships and generate the right answer.
    Other existing prompting method including image-based \textcolor{cyan}{Set-of-Marks}, \textcolor{ForestGreen}{3DAxies} and text-based \textcolor{darkgray}{Chain-of-Thought} failed to answer correctly.}
    %\vspace{-0.1cm}
    \label{fig:case_study}
    \vspace{-7pt}
\end{figure*}

\noindent \textbf{How does \method differ from other visual prompting methods?} Our proposed method calculates and highlights correspondences between images, aiming to elicit 3D and temporal understanding.
Other visual prompting methods (namely Set-of-Mark~\citep{som}, 3DAxiesPrompts~\citep{3dap}, and Chain-of-thought~\citep{wei2023chainofthought}) can also be viewed as alternative prompting methods. 
Given that the ground-truth answers in existing benchmarks are relatively brief, we selected a scene from ScanQA and manually designed a new question. We qualitatively compare \method against other prompting methods on this new question, as shown in~\Cref{fig:case_study}.

The orange part of~\Cref{fig:case_study} shows our \textcolor{orange}{Coarse Correspondence} labels are recognized by GPT-4V. 
The output answer provides evidence that our coarse correspondence helps GPT-4V develop a mental 3D model of the scene.
\textcolor{cyan}{Set-of-Marks} provides no spatial corresponding information and therefore is unhelpful.
The Axis labels in \textcolor{ForestGreen}{3DAxies} can be easily misrecognized by GPT-4V, leading to misleading spatial information.
Though \textcolor{darkgray}{Chain-of-Thought} helps
identify objects, it fails to resolve the ``spatial perspective-taking'' issue.

\begin{figure*}
    \centering
    \includegraphics[width=\linewidth]{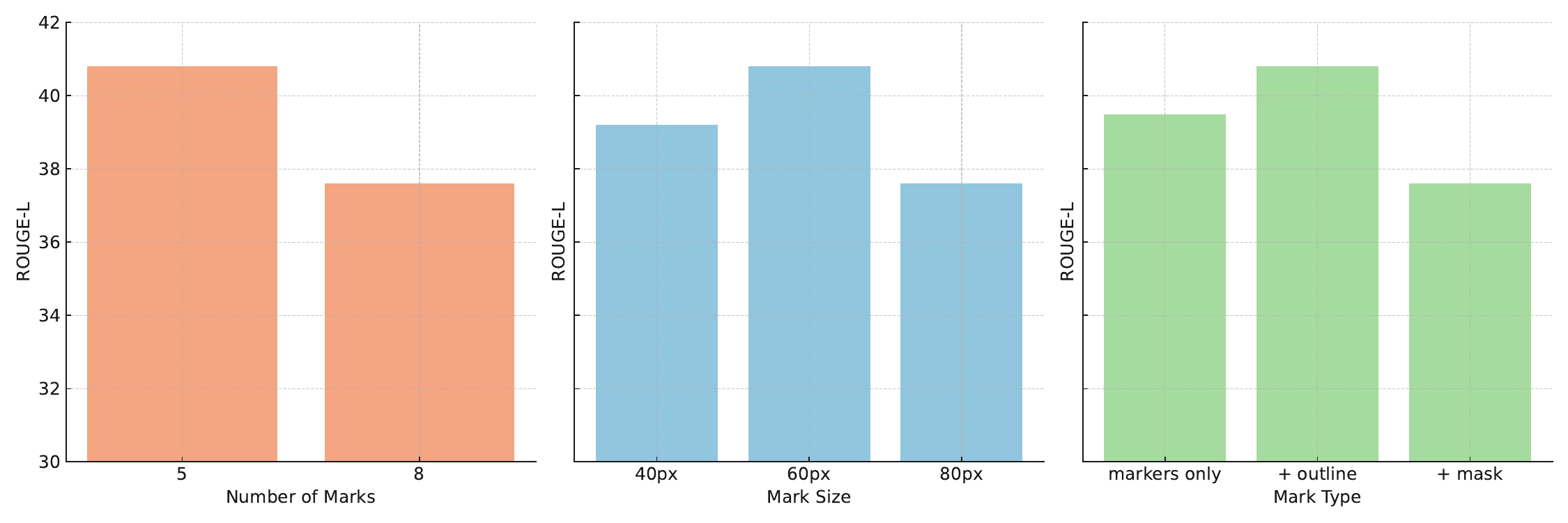}
    %\vspace{-20pt}
    \caption{\textbf{Ablations on different design choices of \method. } We studied the impact of the number, size, and type of marks on performance. All experiments were conducted on ScanQA using GPT-4V.}
    %\vspace{-0.1cm}
    \label{fig:ablation}
    \vspace{-7pt}
\end{figure*}

\noindent \textbf{Why use coarse instead of dense correspondences?}
%ablation on number of selected 
Instead of filtering and retaining only a handful of coarse correspondences, one ablation we considered is the possibility of using all dense correspondence. Unfortunately, we find that excessively overlaying too many instance marks can degrade performance (Figure~\ref{fig:ablation}) as they occlude the visual content in the images.

\noindent \textbf{How large should the marks be?}
We inject the correspondences into MLLMs by overlaying the marks into images. 
We empirically find an optimal mark size (where `px' represents the mark's diameter in pixels) in Figure~\ref{fig:ablation}. Marks that are too small tend to be ignored while those that are too large occlude visual content.
% Balancing the size to effectively convey correspondence information with maintaining the comprehension of contextual semantic information is crucial.

\noindent \textbf{What shape should the marks be?}
We further studied the appearance of the marks. In addition to red circles with white text, we experimented with adding segmentation outlines and segmentation masks. As shown in Figure~\ref{fig:ablation}, using segmentation outlines enhances object grounding. However, using segmentation marks occludes visual content and reduces performance.

\noindent \textbf{Improvement of Camera Motion Invariance with \method.}
A crucial aspect of strong spatial reasoning is the model’s ability to maintain an understanding of spatial relationships invariant to camera motion; in other words, whether the image sequence is captured from left-to-right or right-to-left, the model’s comprehension of static space should remain consistent. However, current benchmarks lack annotations regarding whether the 3D space scan was conducted from left-to-right or right-to-left, making it challenging to analyze the effect of camera motion on MLLMs' spatial understanding. To address this, we curated a small diagnostic benchmark specifically for this analysis. This benchmark includes 10 scenes captured from left-to-right, with five questions per scene focusing on left-right spatial relationships between objects—fundamental aspects of spatial understanding. We ensured that all questions required information across multiple frames and could not be answered from a single frame alone.
Ideally, regardless of whether the image sequence is input in the original or reversed order, the output should remain consistent. We leave additional details on data curation in the supplementary material. 

To mitigate randomness, each question was tested 20 times, yielding a total of 1000 trials, and we report the average accuracy in Table~\ref{tab:realsot}. We then reversed the sequence order to simulate a right-to-left capture and repeated 1000 trials with this reversed input, reporting the average accuracy for this reverse sequence. Finally, we report the harmonic mean of the average accuracies for the forward and reverse inputs.

Without \method, the accuracy in understanding spatial relationships in right-to-left sequences was significantly lower than in left-to-right sequences. With \method, not only did MLLMs achieve a substantial increase in spatial understanding accuracy with the original sequence (+13.2\%), but more importantly, reversing the input order no longer drastically impacted the model's comprehension of static spatial relationships. Specifically, \method improved the harmonic mean of forward and reverse sequence accuracy by 17.3\%. This demonstrates that our method enables MLLMs to perform spatial reasoning invariant to camera motion, thus achieving more robust spatial understanding capabilities.

\begin{table}
  \centering
  \adjustbox{width=\linewidth}{
    \begin{tabular}{c|c|c c c}
      \toprule 
      Models & Frame & Forward & Reverse & Harmonic Mean \\
      %\midrule
      %GPT-4O & 2 & 58.2 & 50.0 & 53.8 \\
      %\rowcolor{gray!20} GPT-4O+CC & 2 & \textbf{71.6} & \textbf{70.6} & \textbf{71.1} \\
      \midrule
      GPT-4O & 4 & 58.0 & 50.4 & 53.9 \\
      \rowcolor{gray!20} GPT-4O+CC & 4 & \textbf{71.2} & \textbf{71.2} & \textbf{71.2} \\
      \bottomrule
    \end{tabular}
  }
  \caption{\textbf{Effect of \method on Camera Motion Bias.} \method shows strong capability of enhancing 3D spatial understanding of MLLMs. It can also ease the camera motion bias of current MLLMs.}
  \label{tab:realsot}
  \vspace{-10pt}
\end{table}

\section{Related work}
\label{sec:related}

\noindent \textbf{Multimodal language models}
Multimodal LLMs \citep{liu2024visual,bai2023qwenvl} integrate vision encoders~\citep{radford2021learning} into large LLMs~\citep{vicuna2023,touvron2023llama}, enabling direct reasoning over visual input. Proprietary models like GPT-4~\citep{OpenAI_GPT4V}, Gemini~\citep{geminiteam2024gemini}, and Claude~\citep{anthropic2024claude}, along with open-source models such as the LLaVA series~\citep{liu2024visual} and BLIP series~\citep{blip2}, have made significant strides in 2D vision-language tasks like image captioning~\citep{chen2015microsoft} and visual question answering (VQA)~\citep{hudson2019gqa,goyal2017making}.
However, for applications in real-world scenarios like autonomous driving~\citep{tian2024drivevlm} and robotics~\citep{yang2023octopus}, MLLMs still require stronger spatial-temporal reasoning capabilities. Recent work~\citep{lin2023video} has enhanced MLLMs' capacity to process multiple image frames and understand temporal dynamics, achieving notable improvements in video dense captioning~\citep{krishna2017densecaptioning} and videoQA~\citep{xiao2021nextqanext,grunde2021agqa}. Nonetheless, MLLMs still face challenges in understanding complex 3D spatial relations~\citep{OpenEQA2023} and long-term temporal dynamics~\citep{mangalam2023egoschema} present in image sequences. Many approaches have introduced specialized architectures~\citep{3d-llm} or fine-tuning methods targeting spatial and temporal reasoning individually. Our goal, however, is to jointly enhance spatial-temporal reasoning within the existing general-purpose MLLM framework. By leveraging off-the-shelf tracking models to extract instance-level correspondences, we achieve this in a training-free manner.

\noindent \textbf{Visual prompting.}
Effective prompting has been widely proven to improve LLMs across multiple domains. Methods, such as chain-of-thought prompting~\citep{wei2023chainofthought}, force the model to reason before answering a question.
In the context of visual prompting, we can enhance MLLMs' grounding abilities~\citep{red-circle,som}, their ability to use numerical expressions to describe 3D object relationships~\citep{3dap}, and their capability to generate outputs for robotics control~\citep{nasiriany2024pivot}, all using single-image inputs.
Unlike these approaches, \method is designed to enhance MLLMs’ spatial-temporal reasoning capabilities for interpreting complex spacetime structures represented in image sequences.
\section{Conclusion.}
We propose a framework called \method prompting. By using off-the-shelf video tracking models to obtain class-agnostic, instance-level correspondences and conveying this information to MLLMs through visual prompting, we discovered that this simple method, using only 2D images as input—without any specialized architectural design or task-specific SFT—can effectively enhance MLLMs’ spatial-temporal reasoning. This improvement extends to embodied tasks like navigation. Our method not only works on proprietary models but also generalizes to open-source models, and it performs well on both in-domain and out-of-domain datasets. Moreover, it enhances not just inference but also training. Further analysis shows that our method helps MLLMs become more robust to camera motion bias.
\section*{Acknowledgement}
We thank Xiaojuan Wang, Ruitao Zhang, Yuntian Deng for helpful discussions, feedback and collecting data. This project is partially funded by Amazon Science. 
{
    \small
    \bibliographystyle{ieeenat_fullname}
    \bibliography{main}
}

\end{document}